\documentclass[journal,twocolumn]{IEEEtran}

\usepackage{xcolor,soul,framed} 
\colorlet{shadecolor}{yellow}
\usepackage[pdftex]{graphicx}
\graphicspath{{../pdf/}{../jpeg/}}
\DeclareGraphicsExtensions{.pdf,.jpeg,.png}

\usepackage[cmex10]{amsmath} 
\usepackage{amssymb} 
\usepackage{array}
\usepackage{cite}
\usepackage{mdwmath}
\usepackage{mdwtab}
\usepackage{eqparbox}
\usepackage{url}
\usepackage{float}
\usepackage{algorithm}
\usepackage{algpseudocode} 
\usepackage{lipsum} 

\usepackage{booktabs} 
\usepackage{tabularx}
\usepackage{caption}
\usepackage{multirow}
\usepackage{hyperref}
\usepackage{pifont}
\usepackage{fancyhdr} 
\usepackage{xcolor}
\usepackage{subcaption}
\usepackage{algorithm}
\hypersetup{hidelinks}
\usepackage{mathrsfs}

\begin{document}

\title{A Modality-Aware Cooperative Co-Evolutionary Framework for Multimodal Graph Neural Architecture Search}

\author{Sixuan~Wang,
        Jiao~Yin,~\IEEEmembership{Member,~IEEE},
        Jinli~Cao,~\IEEEmembership{Member,~IEEE},
        Mingjian~Tang,
        Yong-Feng~Ge,~\IEEEmembership{Member,~IEEE}%
\thanks{Sixuan Wang, Jinli Cao, and Mingjian Tang are with the Department of Computer Science and Information Technology, La Trobe University, Melbourne, VIC 3086, Australia (e-mails: \{Sixuan.Wang, j.cao, Ming.Tang\}@latrobe.edu.au).}%
\thanks{Jiao Yin and Yong-Feng Ge are with the Institute for Sustainable Industries and Liveable Cities, Victoria University, Melbourne, VIC 3011, Australia (e-mails: \{jiao.yin, yongfeng.ge\}@vu.edu.au).}%

\thanks{This work was supported in part by the Australian Research Council under Discovery Project Grant No. DP230100716.}%

\thanks{This work has been submitted to the IEEE for possible publication. Copyright may be transferred without notice, after which this version may no longer be accessible.}%
}


\markboth{IEEE Transactions on Cybernetics,~Vol.~XX, No.~XX, Month~2025}%
{Author \MakeLowercase{\textit{et al.}}: A Modality Aware Cooperative Co-Evolutionary Framework for MGNAS}

\maketitle

\begin{abstract}
Co-exploitation attacks on software vulnerabilities pose severe risks to enterprises, a threat that can be mitigated by analyzing heterogeneous and multimodal vulnerability data. Multimodal graph neural networks (MGNNs) are well-suited to integrate complementary signals across modalities, thereby improving attack-prediction accuracy. However, designing an effective MGNN architecture is challenging because it requires coordinating modality-specific components at each layer, which is infeasible through manual tuning. Genetic algorithm (GA)-based graph neural architecture search (GNAS) provides a natural solution, yet existing methods are confined to single modalities and overlook modality heterogeneity. To address this limitation, we propose a modality-aware cooperative co-evolutionary algorithm for multimodal graph neural architecture search, termed MACC-MGNAS. First, we develop a modality-aware cooperative co-evolution (MACC) framework under a divide-and-conquer paradigm: a coordinator partitions a global chromosome population into modality-specific gene groups, local workers evolve them independently, and the coordinator reassembles chromosomes for joint evaluation. This framework effectively captures modality heterogeneity ignored by single-modality GNAS. Second, we introduce a modality-aware dual-track surrogate (MADTS) method to reduce evaluation cost and accelerate local gene evolution. Third, we design a similarity-based population diversity indicator (SPDI) strategy to adaptively balance exploration and exploitation, thereby accelerating convergence and avoiding local optima. On a standard vulnerabilities co-exploitation (VulCE) dataset, MACC-MGNAS achieves an F1-score of 81.67\% within only 3 GPU-hours, outperforming the state-of-the-art competitor by 8.7\% F1 while reducing computation cost by 27\%.
\end{abstract}


\IEEEpeerreviewmaketitle

\section{Introduction}
\label{sec:intro}

Cyberattacks increasingly exploit multiple vulnerabilities in concert (vulnerability co-exploitation)~\cite{lu2025evolutionary}, enabling rapid privilege escalation, lateral propagation, and persistent footholds for data theft and service disruption. Public reports indicate approximately 133 daily CVE disclosures, with over 20\% involving co-exploitation chains~\cite{ghosh2025cve}. 

Analyzing software vulnerability data—particularly signals from textual descriptions, severity scores, and dependency relations—is an effective approach to mitigating co-exploitation risk. For instance, the traditional machine-learning model, \emph{Common Vulnerability Scoring System–Bidirectional Encoder Representations from Transformers (CVSS-BERT)}~\cite{shahid2021cvss}, derives severity vectors solely from text. However, single-modality features capture only part of the available evidence and underrepresent real-world complexity, thereby limiting predictive performance. Recent studies address this issue by integrating multimodal signals into \emph{Vulnerability Knowledge Graphs (VulKGs)}~\cite{yin2024compact}, where the subgraph focusing on co-exploitation relations is denoted as \emph{Vulnerabilities Co-Exploitation (VulCE)}. Building on VulCE, Yin et al. proposed \emph{a modality-aware graph convolutional network (MAGCN)} within the multimodal graph neural network (MGNN) family and selected its core components through exhaustive grid search, which improved prediction performance. However, grid search across multimodal design choices is computationally prohibitive and does not scale, thereby motivating automated design via graph neural architecture search (GNAS).

Given the inefficiency of manual design, GNAS automates MGNN construction by searching over message functions, aggregation and update operators, and cross-modal fusion mechanisms. Exhaustive strategies such as grid search are infeasible~\cite{yin2022knowledge}, while reinforcement learning- or gradient-based methods, though explored for graph models, are computationally expensive due to their reliance on gradient information and are ill-suited to the highly discrete MGNN design space~\cite{salmani2025systematic}. In contrast, genetic algorithms (GAs) are well suited for this task because they do not depend on gradient information~\cite{wang2025abg}. Moreover, GA-based approaches naturally encode MGNN architectures as chromosomes, where diverse structural components form a population of candidate solutions. Candidate quality can then be estimated efficiently through techniques such as early-stopped training, surrogate-assisted evaluation, and parallelization~\cite{ma2025two}. For example, the \emph{Divide-and-Conquer Neural Architecture Search (DC-NAS)} applies cooperative co-evolution in a two-level GA framework to optimize fusion modules~\cite{liang2024dc}.

Despite these advances, designing MGNNs for VulCE-based co-exploitation prediction remains particularly challenging. First, structural heterogeneity dramatically enlarges the design space, complicating coordination among modality-specific components such as message functions, fusion operators, and propagation layers. Consequently, manual designs frequently result in suboptimal architectures. Second, although GA-based GNAS provides a promising direction for automated MGNN design, existing approaches suffer from three major limitations: (i) most overlook modality-specific characteristics, (ii) cooperative co-evolutionary frameworks incur prohibitive computational costs due to the evaluation of numerous candidate architectures in each generation, and (iii) conventional implementations lack adaptive mechanisms to balance exploration and exploitation, leading to slow convergence or premature convergence to suboptimal solutions.

To address these challenges, this work proposes \emph{MACC-MGNAS}, a modality-aware cooperative co-evolutionary algorithm for MGNN architecture search. It guides design at multiple granularities and explicitly incorporates modality-specific characteristics. The main contributions of this work are as follows:

\begin{enumerate}
\item \textbf{Formulation of MGNN Architecture as a GNAS Problem}: This work is the first to cast full MGNN architecture design as a discrete, multimodal GNAS problem, with a search space covering core components (message functions, aggregation and update operators, and cross-modal fusion). This formulation enables systematic comparison of design choices and reduces reliance on extensive expert tuning. 

\item \textbf{MACC: Divide-and-Conquer Cooperative Co-Evolution Framework}: A modality-aware cooperative co-evolution framework, termed MACC, is introduced under a divide-and-conquer paradigm. A coordinator performs global evolution over chromosome-level architectures, while workers conduct local evolution on modality-specific and fusion gene blocks. This design enables coordinated search across global and local scopes, preserves modality-specific characteristics, and reduces combinatorial complexity. 

\item \textbf{MADTS: Dual-Track Surrogate for Efficient Fitness Estimation}: A modality-aware dual-track surrogate (MADTS) is developed to reduce evaluation cost and accelerate worker-side evolution. Candidate fitness is approximated with two lightweight surrogates (for modality-specific and fusion gene blocks), thereby reducing the number of full-model evaluations and accelerating convergence. 

\item \textbf{SPDI: Diversity Indicator for Adaptive Exploration–Exploitation Balance}: A similarity-based population diversity indicator (SPDI) is proposed to quantify diversity in the coordinator-side search and to adaptively balance exploration and exploitation by adjusting crossover and mutation rates. This strategy accelerates convergence while mitigating premature convergence and avoiding local optima. 
\end{enumerate}

The remainder of this paper is organized as follows. Section~\ref{sec:RL} reviews related work. Section~\ref{sec:problem} defines the MGNN search space and formulates the GNAS problem. Section~\ref{sec:methodology} presents the proposed algorithm. Section~\ref{experiment and analysis} reports experimental results and analysis. Section~\ref{conclusion} concludes the paper and discusses future directions.

\section{Related Work}
\label{sec:RL}

\subsection{MGNN Architecture Design: Manual vs. GNAS}
\label{sec:manual_vs_nas}

Early MGNNs relied on handcrafted designs, where each modality was encoded and fused through manually specified operators. For instance, \emph{Vulnerability Detection based on Multi-Dimensional Similar Neighbors (VulSim)}~\cite{shimmi2024vulsim} incorporates multi-dimensional neighbor embeddings for vulnerability detection, but its fusion scheme is fixed and cannot adapt to other tasks. To capture richer cross-modal interactions, the \emph{Interactive Multimodal Fusion Model (IMF)}~\cite{li2023imf} applies Tucker decomposition-based bilinear pooling, but this flexibility requires extensive expert tuning of fusion operators, limiting scalability across tasks. Similarly, the \emph{Edges Graph Neural Network (EGNN)}~\cite{zhang2024edge} introduces edge-specific propagation to model relation-aware interactions, but such tailored designs struggle to scale when inputs become heterogeneous or modalities proliferate. More recent variants, such as \emph{ModalitY Information as Fine-Grained Tokens (MYGO)}~\cite{zhang2025tokenization} and \emph{Cross-Modal Consistency and Relation Semantics (C2RS)}~\cite{shu2025c}, attempt to overcome these issues by enforcing cross-modal consistency or introducing modality tokenization. Nevertheless, handcrafted MGNNs remain task-specific and dependent on expert-crafted architectures, fundamentally restricting their scalability.

To mitigate this reliance on manual heuristics, researchers explored automated strategies. A representative early attempt is \emph{MAGCN}~\cite{yin2022knowledge}, which employed grid search over MGNN components but quickly became infeasible as the design space expanded. This motivated the adoption of GNAS. Early studies investigated alternative search paradigms. Particle swarm optimization (PSO)~\cite{huang2023split} converges quickly but often yields suboptimal architectures. Bayesian optimization (BO)~\cite{zhang2023surrogated} probabilistically models the search space and achieves stronger performance, but its sequential nature limits scalability. Estimation of distribution algorithms (EDAs)~\cite{li2023surrogate} and GAs~\cite{wang2025abg} introduced population-based strategies; however, the highly discrete MGNN space reduces the effectiveness of EDAs, whereas GAs naturally capture diverse structural patterns through crossover and mutation, making them particularly effective.

Building on GA-based search, subsequent frameworks attempted to address multimodality from different perspectives. \emph{Bilevel Multimodal Neural Architecture Search (BM-NAS)}~\cite{yin2022bm} introduces bilevel optimization to decouple fusion and propagation, but at the cost of additional overhead and without explicit modality adaptation. \emph{Out-of-Distribution Generalized Multimodal Graph Neural Architecture Search (OMG-NAS)}~\cite{cai2024omgnas} improves robustness by reducing distribution shift, yet modality heterogeneity remains largely overlooked. \emph{DC-NAS}~\cite{liang2024dc} adopts cooperative co-evolution to enhance efficiency, but it was tailored only to fusion design rather than the full MGNN architecture.

Overall, while these approaches mark significant progress, they remain constrained by scalability issues and the lack of modality-aware coordination. These limitations directly motivate the development of a cooperative co-evolutionary framework that explicitly accounts for modality-specific characteristics in MGNN design, as pursued in this work.

\subsection{Surrogate Models for Efficient Evaluation in GNAS}
\label{sec:surrogate_models}

A major bottleneck in GA-based GNAS is the prohibitive cost of evaluating candidate architectures, as each fitness assessment typically requires full training and validation~\cite{li2023surrogate,wang2025abg}. Surrogate models alleviate this by approximating performance, thereby reducing the number of expensive training cycles. While transferable surrogates~\cite{qin2025transferrable} and zero-cost proxies~\cite{qiao2024tg} improve efficiency in GNAS, they primarily target unimodal spaces and fail to capture the complexity of multimodal fusion operators. 

Gaussian process (GP) surrogates provide a stronger alternative. For instance, GP-NAS~\cite{li2020gp} and Li et al.~\cite{li2023gaussian} showed that GPs offer sample-efficient modeling of discrete architecture spaces, making them particularly attractive for GNAS. Building on this, Wang et al.~\cite{wang2025knowledge} introduced a knowledge-aware GP surrogate that leverages prior architectures and datasets for faster cold-start optimization. While effective in reducing warm-up costs, this approach still assumes relatively homogeneous search spaces and does not explicitly address modality-specific structures. To further enhance scalability, Gong et al.~\cite{gong2024offline} proposed the \emph{Cooperative Coevolution-Based Data-Driven Evolutionary Algorithm (CC-DDEA)}, which integrates offline GP surrogates within a hierarchical surrogate–joint learning model. However, CC-DDEA likewise neglects multimodal heterogeneity, particularly in fusion mechanisms, limiting its effectiveness for MGNNs.

In summary, surrogate models have significantly advanced the efficiency of GNAS, but existing approaches largely overlook multimodal heterogeneity. This gap motivates our modality-aware surrogate strategy, which explicitly captures modality dependencies while preserving scalability in MGNN search.

\section{Problem Formulation}
\label{sec:problem}

\subsection{Definition of MGNN Architecture}
\label{sec:mgnn_definition}

We formalize a multimodal graph as $\mathcal{G} = \big(\mathcal{V}, \{\mathcal{E}^{(m)}\}_{m=1}^{M}, \{\mathbf{X}^{(m)}\}_{m=1}^{M}\big)$, where $\mathcal{V}$ is the node set, $\mathcal{E}^{(m)}$ and $\mathbf{X}^{(m)}$ are, respectively, the modality–specific edges and node features for modality $m$; throughout, the superscript $(m)$ denotes the $m$-th modality ($m=1,\dots,M$). Each node $i\in\mathcal{V}$ carries a hidden representation $\mathbf{h}_i$. For a neighbor $j\in\mathcal{N}_i^{(m)}$ in modality $m$, a message is computed as $\mathbf{m}_{i,j}^{(m)}=\phi^{(m)}(\mathbf{h}_i,\mathbf{h}_j,\mathbf{x}_{ij}^{(m)})$ and then aggregated within the modality as $\mathbf{a}_i^{(m)}=\psi^{(m)}\!\big(\{\mathbf{m}_{i,j}^{(m)}: j\in\mathcal{N}_i^{(m)}\}\big)$. The node state is updated using all modalities, $\mathbf{h}_i'=\gamma\!\big(\mathbf{h}_i,\{\mathbf{a}_i^{(m)}\}_{m=1}^{M}\big)$, and a cross-modal fusion operator integrates modality–wise representations into a unified embedding $\mathbf{z}_i=\mathcal{F}\!\big(\mathbf{a}_i^{(1)},\dots,\mathbf{a}_i^{(M)}\big)$. A readout head then maps node- or graph-level embeddings to task outputs, $\hat{y}=\mathcal{R}\big(\{\mathbf{z}_i\}_{i\in\mathcal{V}}\big)$. The propagation depth $L$ denotes the number of stacked MGNN layers (i.e., the receptive-field radius).

Thus, the MGNN can be summarized by the architecture tuple in Eq.~\eqref{eq:architecture}:
\begin{equation}
A = (\phi^{(m)}, \psi^{(m)}, \gamma, \mathcal{F}, \mathcal{R}, L),
\label{eq:architecture}
\end{equation}
where each component is selected from a finite candidate set.

\subsection{Formulating MGNN Design as a GA-based GNAS Problem}
\label{sec:ga_formulation}

Based on Eq.~\eqref{eq:architecture}, we encode each MGNN $A$ as a chromosome $\mathbf{C}$ to enable GA-based search.  
Each chromosome $\mathbf{C}$ corresponds uniquely to one architecture $A$, and vice versa.

\paragraph{Chromosome Encoding.} 
The encoding is defined in Eq.~\eqref{eq:chromosome}, where each gene corresponds to one architectural component:
\begin{equation}
\mathbf{C} = (g_1, g_2, \ldots, g_K),
\label{eq:chromosome}
\end{equation}
with $g_i$ chosen from the MGNN components in Eq.~\eqref{eq:architecture}.  
For instance, one gene may correspond to the design choice of a modality-specific message-passing component, while another may represent the selection of a cross-modal fusion operator. This chromosome-level representation enables MGNN architectures to be systematically encoded and manipulated as individuals in a GA population.

\paragraph{Search Space.} 
Let $\mathcal{S}$ denote the discrete chromosome space induced by all valid encodings. The size of $\mathcal{S}$ grows combinatorially with the number of modalities $M$ and component choices, making exhaustive enumeration infeasible.  

\paragraph{Optimization Objective and Complexity.} 
The goal is to identify the best-performing chromosome, as formulated in Eq.~\eqref{eq:gnas_opt}:
\begin{equation}
\mathbf{C}^{*} = \arg\max_{\mathbf{C} \in \mathcal{S}} f(\mathbf{C}),
\label{eq:gnas_opt}
\end{equation}
where $f(\mathbf{C})$ is the fitness function (e.g., validation accuracy), possibly approximated by surrogates.  
Eq.~\eqref{eq:gnas_opt} is equivalent in complexity to a multi-choice knapsack problem (MCKP)~\cite{duan2025adaptable}, which makes MGNN architecture search NP-hard~\cite{liu2024combinatorial}.  

\paragraph{GA-based Search.} 
To address this hardness, GAs evolve a population of chromosomes $\{\mathbf{C}_i\}_{i=1}^P$ through selection, crossover, and mutation. Selection preserves high-quality designs, while crossover and mutation introduce diversity. During evolution, $f(\mathbf{C})$ can be estimated via partial training or surrogate models, which not only makes the search scalable but also naturally balances exploration and exploitation. This population-based strategy is particularly well suited to MGNNs, as it enables coordinated exploration of modality-specific components and fusion strategies under limited computational budgets.

\section{Methodology}
\label{sec:methodology}

\subsection{Overview of Proposed MACC-MGNAS Algorithm}
\label{sec:overview}

\begin{figure}[ht]
\centering
\includegraphics[width=1\linewidth]{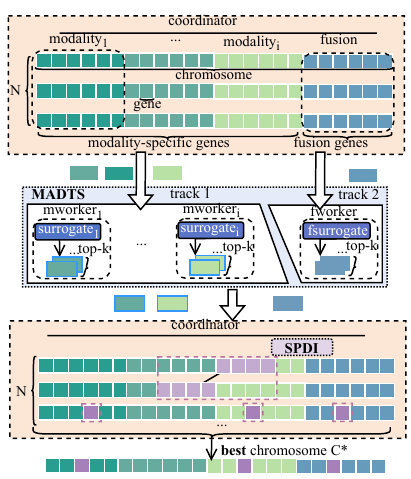}
\caption{Schematic structure of the MACC-MGNAS algorithm. 
A global coordinator interacts with multiple modality workers (mworkers) 
and one fusion worker (fworker). 
The coordinator decomposes chromosomes into modality-specific and fusion blocks, 
dispatches them to workers for local optimization, and then collects feedback 
to update the global population.}
\label{fig:framework}
\end{figure}

As illustrated in Fig.~\ref{fig:framework}, the overall framework of MACC-MGNAS adopts a divide-and-conquer strategy under the \textbf{MACC framework} (Sec.~\ref{sec:macc}), where a global \emph{coordinator}) interacts with multiple local \emph{workers} (middle of Fig.~\ref{fig:framework}). 

At the coordinator side, each MGNN architecture is encoded as a chromosome that combines modality-specific genes and fusion genes. The coordinator decomposes chromosomes into these two groups and dispatches them to the corresponding workers for localized optimization.  

At the worker side, modality workers (\emph{mworkers}) and the fusion worker (\emph{fworker}) independently optimize their assigned gene subsets. To reduce evaluation cost, the \emph{MADTS} module (Sec.~\ref{sec:dts}) employs surrogate models tailored to modality and fusion genes, enabling efficient and parallel fitness estimation. Each worker then returns its top-performing candidates to the coordinator.  

The coordinator then recombines the locally optimized genes into complete chromosomes for global evolution. To further prevent premature convergence and enhance exploration, the \emph{SPDI} mechanism (Sec.~\ref{sec:SPDI}) adaptively adjusts crossover and mutation rates during global search. Through repeated decomposition, local surrogate optimization, and global recombination, MACC-MGNAS iteratively refines the population and identifies the best-performing architecture $C^*$.

\subsection{MACC Framework}
\label{sec:macc}

\begin{figure}[h]
\centering
\includegraphics[width=1\linewidth]{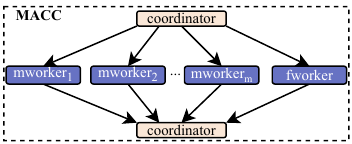}
\caption{Illustration of the proposed MACC framework. 
Each chromosome is decomposed into modality-specific and fusion blocks (Eq.~\eqref{eq:macc-decomposition}), 
which are independently optimized by modality workers (mworkers) and a fusion worker (fworker). 
The elite blocks returned by workers are merged (Eq.~\eqref{eq:macc-merge}) at the Coordinator side to construct candidate chromosomes, which are evaluated at the global level (Eq.~\eqref{eq:fitness-global}).}
\label{fig:macc}
\end{figure}

To address the challenges of MGNN architecture search, we design the \emph{MACC} framework under a divide-and-conquer paradigm. 
MACC organizes GA-based GNAS into two interacting levels: 
a \textit{global coordinator} that evolves chromosomes representing complete architectures,
and multiple \textit{workers} (\emph{mworkers} for modality-specific blocks and one \emph{fworker} for fusion) 
that refine subsets of genes in parallel. 
This design reduces the combinatorial complexity of the search space,
preserves modality-specific characteristics, and enables coordinated global optimization. 
The overall process for one global generation is illustrated in Fig.~\ref{fig:macc}.

\paragraph{Chromosome Decomposition.}
As defined in Eq.~\eqref{eq:chromosome}, each chromosome $\mathbf{C}$ encodes an MGNN architecture as a sequence of $K$ genes. 
In MACC, $\mathbf{C}$ is decomposed into modality-specific and fusion blocks, as given by Eq.~\eqref{eq:macc-decomposition}:
\begin{equation}
\mathbf{C} = \big( \mathbf{C}^{(1)}, \mathbf{C}^{(2)}, \ldots, \mathbf{C}^{(M)}, \mathbf{C}^{(\text{fus})} \big),
\label{eq:macc-decomposition}
\end{equation}
where $\mathbf{C}^{(m)}$ denotes the block assigned to the $m$-th modality worker 
and $\mathbf{C}^{(\text{fus})}$ denotes the block handled by the fusion worker.  

\paragraph{Coordinator and Workers.}  
The \textbf{Coordinator} maintains the global population of chromosomes, defined in Eq.~\eqref{eq:population}:
\begin{equation}
\mathcal{P}_t = \{\mathbf{C}_1, \mathbf{C}_2, \ldots, \mathbf{C}_N\}, 
\label{eq:population}
\end{equation}
where $N$ is the population size and each $\mathbf{C}_i$ encodes a candidate MGNN architecture.  
At each generation $t$, the coordinator decomposes chromosomes in $\mathcal{P}_t$ (Eq.~\eqref{eq:macc-decomposition}) and dispatches the blocks to workers.  
On the worker side, mworkers and the fworker optimize their assigned blocks through genetic operations.  
After $T_{LS}$ local steps, each worker reports an elite set of blocks, denoted by $E^{(m)}_t$ and $E^{(\text{fus})}_t$, back to the coordinator.  
The coordinator then merges these elites to form candidate chromosomes (Eq.~\eqref{eq:macc-merge}):  
\begin{equation}
\widetilde{\mathcal{P}}^{(t)} = 
\Big(\prod_{m=1}^{M} E^{(m)}_t\Big) \times E^{(\text{fus})}_t,
\label{eq:macc-merge}
\end{equation}
where $\widetilde{\mathcal{P}}^{(t)}$ is the set of candidates constructed from the elite blocks.  

\paragraph{Fitness Evaluation.}
Consistent with Eq.~\eqref{eq:gnas_opt}, 
the global fitness of a chromosome $\mathbf{C}$ is given by Eq.~\eqref{eq:fitness-global}:
\begin{equation}
F_{\mathrm{global}}(\mathbf{C}) 
= \mathrm{Eval}\big(\mathcal{M}(\mathbf{C}); \mathcal{D}_{val}\big),
\label{eq:fitness-global}
\end{equation}
where $\mathcal{M}(\mathbf{C})$ is the MGNN instantiated from chromosome $\mathbf{C}$ 
and $\mathcal{D}_{val}$ is the validation dataset.  

At the local level, each worker computes auxiliary fitness (Eq.~\eqref{eq:fitness-local}):
\begin{equation}
F^{(m)}_{\mathrm{local}}(\mathbf{C}^{(m)}) 
= \mathrm{ModuleEval}\big(\mathbf{C}^{(m)}; \mathcal{D}_{val}^{(m)}\big),
\label{eq:fitness-local}
\end{equation}
where $\mathcal{D}_{val}^{(m)}$ is the validation subset for modality $m$.  
Local fitness (Eq.~\eqref{eq:fitness-local}) is used to update surrogates and guide worker-side evolution, 
while the optimization objective remains maximizing the global fitness (Eq.~\eqref{eq:fitness-global}).

\paragraph{Co-Evolution Procedure.}
The cooperative search proceeds for $T$ global generations, following a collaborative GA evolution cycle. 
At each generation $t$, each chromosome $\mathbf{C}$ (encoding an MGNN architecture as defined in Eq.~\eqref{eq:chromosome}) 
is decomposed into modality-specific and fusion blocks (Eq.~\eqref{eq:macc-decomposition}) 
and dispatched by the \textit{Coordinator} to the \textit{Workers}. 
The mworkers and fworker independently evolve their assigned blocks for $T_{LS}$ local steps using GA operators 
(selection, crossover, and mutation), and return elite blocks. 
The \textit{Coordinator} then cooperatively merges these elites (Eq.~\eqref{eq:macc-merge}) to reconstruct candidate chromosomes. 
Each chromosome is subsequently \emph{decoded} into an instantiated MGNN $\mathcal{M}(\mathbf{C})$ 
and evaluated for global fitness (Eq.~\eqref{eq:fitness-global}). 
Through this iterative encoding–decomposition–local evolution–merging–decoding cycle, 
the \textit{Coordinator} and \textit{Workers} jointly perform a GA-based co-evolutionary process 
that balances global exploration with modality-specific adaptation. 
After $T$ generations, the best chromosome $\mathbf{C}^*$ is selected and decoded as the final MGNN architecture.

\subsection{MADTS Method}
\label{sec:dts}

To improve the efficiency of GA-based GNAS, we design the \emph{MADTS} method as a dual-track surrogate strategy embedded in the MACC framework. MADTS operates at the worker side to approximate local fitness of modality and fusion blocks, thereby reducing expensive full-model evaluations while retaining the Coordinator’s global evaluation as the ground truth (Fig.~\ref{fig:framework}, middle panel).

\paragraph{Two tracks and archive update.}
For each modality block $\mathbf{C}^{(m)}$, a \textbf{Gaussian process surrogate} maps block encodings to predicted scores, as defined in Eq.~\eqref{eq:dts-surrogate}:
\begin{equation}
\hat{F}^{(m)}(\mathbf{C}^{(m)}) 
= S^{(m)}(\mathbf{C}^{(m)};\,\theta^{(m)}),
\label{eq:dts-surrogate}
\end{equation}
where $S^{(m)}$ is the surrogate regressor with parameters $\theta^{(m)}$. 

For the fusion block $\mathbf{C}^{(\mathrm{fus})}$, no standalone local fitness exists (cf. Eq.~\eqref{eq:fitness-local}). 
Instead, the fusion surrogate generates candidate blocks via randomized sampling, whose quality is assessed only through the global fitness $F_{\mathrm{global}}(\mathbf{C})$ (Eq.~\eqref{eq:fitness-global}).

Each worker maintains a bounded archive of candidate blocks. 
For modality workers, the archive is updated by fusing scores (Eq.~\eqref{eq:dts-fused}), removing duplicates, keeping top-$K$ elites, and retraining surrogates. 
For the fusion worker, archive updates rely solely on Coordinator feedback, without retraining.

\paragraph{Acquisition-based proposal.}
Surrogates guide candidate selection via an acquisition function $\mathscr{U}(\cdot)$. For each worker, the next block is selected by Eq.~\eqref{eq:dts-acquisition}:
\begin{equation}
\mathbf{C}^{(m)}_{\text{next}}
= \arg\max_{\mathbf{C}^{(m)} \in \mathcal{S}^{(m)}} 
\mathscr{U}\!\big(\hat{F}^{(m)}(\mathbf{C}^{(m)})\big),
\label{eq:dts-acquisition}
\end{equation}
where $\mathcal{S}^{(m)}$ is the search space for modality $m$. 
Each worker forwards its top-$K$ proposals to the Coordinator for decoding and global evaluation (Eq.~\eqref{eq:fitness-global}).

\paragraph{Coordinator--Worker interaction.}
At generation $t$, workers compute local scores $F^{(m)}_{\mathrm{local}}(\mathbf{C}^{(m)})$ (Eq.~\eqref{eq:fitness-local}), while the Coordinator provides global feedback $F_{\mathrm{global}}(\mathbf{C})$ (Eq.~\eqref{eq:fitness-global}). These are fused into a variance-aware score (Eq.~\eqref{eq:dts-fused}):
\begin{equation}
\widehat{F}^{(m)}_t(\mathbf{C}^{(m)}) 
= \alpha_t \, F_{\mathrm{global}}(\mathbf{C}) 
+ \big(1-\alpha_t\big)\, F^{(m)}_{\mathrm{local}}(\mathbf{C}^{(m)}),
\label{eq:dts-fused}
\end{equation}
where $\alpha_t$ is set adaptively from the variances of local and global scores (Eq.~\eqref{eq:dts-alpha}):
\begin{equation}
\alpha_t = \frac{\widehat{\sigma}^2_{\mathrm{local},t}}
{\widehat{\sigma}^2_{\mathrm{local},t}+\widehat{\sigma}^2_{\mathrm{global},t}+\varepsilon}.
\label{eq:dts-alpha}
\end{equation}

Overall, \textbf{MADTS} accelerates worker-side evolution by proposing high-quality candidates, while the final decision relies on the Coordinator’s global evaluation (Eq.~\ref{eq:fitness-global}).

\subsection{SPDI Strategy}
\label{sec:SPDI}

To stabilize GA-based GNAS, we design \emph{SPDI} as an adaptive diversity controller embedded in the MACC framework. \emph{SPDI} operates at the Coordinator side and quantifies population diversity from chromosome similarity (Fig.~\ref{fig:framework}, bottom panel). Using this signal, the Coordinator adaptively modulates crossover and mutation rates, preventing both diversity collapse and excessive drift.

\paragraph{Diversity measure.}
For the global population $\mathcal{P}_t = \{\mathbf{C}_1, \ldots, \mathbf{C}_N\}$ at generation $t$, 
the pairwise Euclidean distance in chromosome encoding space is defined in Eq.~\eqref{eq:SPDI_pairwise}:
\begin{equation}
d_{ij} = \big\| \pi(\mathbf{C}_i) - \pi(\mathbf{C}_j) \big\|_2, 
\quad i \neq j,
\label{eq:SPDI_pairwise}
\end{equation}
where $\pi(\mathbf{C})$ denotes the vector encoding of chromosome $\mathbf{C}$.  
The similarity-based population diversity is then computed according to Eq.~\eqref{eq:SPDI_mean}:
\begin{equation}
\mathrm{SPDI}(\mathcal{P}_t) 
= \frac{2}{N(N-1)} \sum_{i<j} d_{ij}.
\label{eq:SPDI_mean}
\end{equation}
\textit{Intuitively, SPDI is higher when chromosomes are more dissimilar, and lower when they converge to similar encodings.}

\paragraph{Adaptive adjustment.}
Based on the diversity measure in Eq.~\eqref{eq:SPDI_mean}, 
crossover and mutation rates are adjusted following 
Eqs.~\eqref{eq:SPDI_crossover_rate}–\eqref{eq:SPDI_mutation_rate}:
\begin{align}
p_{\text{cross}}^{(t)} &=
\begin{cases}
p_{\mathrm{cross}}^{\mathrm{h}}, & \mathrm{SPDI}(\mathcal{P}_t) \ge \tau_t, \\[4pt]
p_{\mathrm{cross}}^{\mathrm{l}}, & \mathrm{SPDI}(\mathcal{P}_t) < \tau_t,
\end{cases}
\label{eq:SPDI_crossover_rate} \\[6pt]
p_{\text{mut}}^{(t)} &=
\begin{cases}
p_{\mathrm{mut}}^{\mathrm{l}}, & \mathrm{SPDI}(\mathcal{P}_t) \ge \tau_t, \\[4pt]
p_{\mathrm{mut}}^{\mathrm{h}}, & \mathrm{SPDI}(\mathcal{P}_t) < \tau_t,
\end{cases}
\label{eq:SPDI_mutation_rate}
\end{align}
where $(p_{\mathrm{cross}}^{\mathrm{h}}, p_{\mathrm{mut}}^{\mathrm{l}})$ corresponds to 
an \emph{exploit} mode with strong recombination but weak mutation, 
while $(p_{\mathrm{cross}}^{\mathrm{l}}, p_{\mathrm{mut}}^{\mathrm{h}})$ corresponds to 
an \emph{explore} mode with stronger mutation to maintain diversity.  
Thus, SPDI serves as a dynamic signal of population similarity, 
which the Coordinator uses to adaptively switch between exploration and exploitation.

\paragraph{Operator design.}
To better fit MGNN architecture design, 
we adapt standard GA operators as follows:

\textit{Cross-modality crossover.}  
Given two parent chromosomes $\mathbf{C}^a$ and $\mathbf{C}^b$, their encodings are specified in Eq.~\eqref{eq:SPDI-parents}:
\begin{equation}
\mathbf{C}^a = (g^a_1, \dots, g^a_m, \dots, g^a_K), 
\quad 
\mathbf{C}^b = (g^b_1, \dots, g^b_{m'}, \dots, g^b_K),
\label{eq:SPDI-parents}
\end{equation}
where $m \neq m'$ denote two distinct modalities.  
The resulting offspring are given by Eq.~\eqref{eq:SPDI-crossover}:
\begin{equation}
\mathbf{C}^{a'} = (g^a_1, \dots, g^b_{m'}, \dots, g^a_K), 
\quad
\mathbf{C}^{b'} = (g^b_1, \dots, g^a_m, \dots, g^b_K).
\label{eq:SPDI-crossover}
\end{equation}

\textit{Global mutation.}  
For a chromosome $\mathbf{C} = (g_1, \ldots, g_K)$, single-gene replacement is performed as follows (see Eq.~\eqref{eq:SPDI-mutation}):
\begin{equation}
\mathbf{C}' = (g_1, \ldots, g_{j-1}, g'_j, g_{j+1}, \ldots, g_K), 
\quad g'_j \neq g_j.
\label{eq:SPDI-mutation}
\end{equation}
Compared with the block-level exchange in Eq.~\eqref{eq:SPDI-crossover}, Eq.~\eqref{eq:SPDI-mutation} specifies a point mutation at position $j$.

Together, Eq.~\eqref{eq:SPDI-crossover} demonstrates block-level crossover, 
while Eq.~\eqref{eq:SPDI-mutation} specifies single-gene mutation. 

Overall, \emph{SPDI} provides a population-level, similarity-based control signal that adaptively calibrates the exploration–exploitation balance, mitigating premature convergence and improving the effectiveness of global search within MACC-MGNAS.

\subsection{Algorithmic Workflow and Complexity}
\label{sec:workflow-complexity}

\paragraph{Workflow summary.}
To integrate the proposed components, we summarize the complete workflow of \emph{MACC-MGNAS}. 
The procedure starts with \textbf{initialization} (Line~1): the Coordinator samples the initial population $\mathcal{P}^{(0)}$ of size $N$, initializes worker archives, and records the incumbent best. 
At each global generation $t$, the Coordinator measures population diversity via SPDI and sets crossover/mutation rates (Line~2). 
It then decomposes each chromosome into modality and fusion blocks and dispatches them to workers (Line~3). 
On the worker side, MADTS updates modality surrogates, fuses local/global signals, proposes candidates via the acquisition rule, evolves local populations for $T_{LS}$ steps, and returns elites (Line~4). 
The Coordinator merges returned elites, decodes the reconstructed chromosomes, evaluates global fitness, and selects survivors (Line~5). 
Variation operators are applied to refresh the population (Line~6), and the incumbent best is updated (Line~7). 
After $T$ generations, the algorithm outputs the best chromosome $\mathbf{C}^*$ and optionally the decoded MGNN model (Line~8).

\begin{algorithm}[t]
\caption{MACC-MGNAS}
\label{alg:macc-mgnas}
\begin{algorithmic}[1]
\Require Search space $\mathcal{S}$ (chromosome; Eq.~\eqref{eq:chromosome}), 
population size $N$ (Eq.~\eqref{eq:population}), 
generations $T$, local steps $T_{LS}$, 
datasets $\mathcal{D}_{train}, \mathcal{D}_{val}$; 
MACC split (Eq.~\eqref{eq:macc-decomposition}); 
fitness (Eqs.~\eqref{eq:fitness-global}, \eqref{eq:fitness-local}); 
SPDI (Eqs.~\eqref{eq:SPDI_pairwise}--\eqref{eq:SPDI_mutation_rate}); 
MADTS (Eqs.~\eqref{eq:dts-surrogate}--\eqref{eq:dts-alpha})
\Ensure Best chromosome $\mathbf{C}^*$ \textit{(optionally decoded model $\mathcal{M}(\mathbf{C}^*)$)}
\State \textbf{Initialization:} sample $\mathcal{P}^{(0)}\!\subset\!\mathcal{S}$; init worker archives; set incumbent $\mathbf{C}^*$ using Eq.~\eqref{eq:fitness-global}
\For{$t=1$ \textbf{to} $T$} \Comment{Global generation $t$}
  \State \textbf{SPDI:} compute diversity (Eqs.~\eqref{eq:SPDI_pairwise}, \eqref{eq:SPDI_mean}); set $p_{\text{cross}}^{(t)}$, $p_{\text{mut}}^{(t)}$ (Eqs.~\eqref{eq:SPDI_crossover_rate}--\eqref{eq:SPDI_mutation_rate})
  \State \textbf{MACC split:} decompose chromosomes and dispatch (Eq.~\eqref{eq:macc-decomposition})
  \State \textbf{MADTS (workers):} update surrogates/fuse signals (Eqs.~\eqref{eq:dts-surrogate}, \eqref{eq:dts-fused}, \eqref{eq:dts-alpha}); propose via acquisition (Eq.~\eqref{eq:dts-acquisition}); evolve $T_{LS}$; return elites
  \State \textbf{Merge \& Eval:} recombine elites (Eq.~\eqref{eq:macc-merge}); evaluate $F_{\mathrm{global}}(\mathbf{C})$ (Eq.~\eqref{eq:fitness-global}); select survivors
  \State \textbf{Variation:} crossover/mutation (Eqs.~\eqref{eq:SPDI-crossover}, \eqref{eq:SPDI-mutation}); form $\mathcal{P}^{(t)}$
  \State Update incumbent $\mathbf{C}^*$ by Eq.~\eqref{eq:fitness-global}
\EndFor
\State \Return $\mathbf{C}^*$ \textit{(and optionally $\mathcal{M}(\mathbf{C}^*)$)}
\end{algorithmic}
\end{algorithm}

\paragraph{Complexity analysis.}
Per generation, the computational cost mainly comes from four sources. 
SPDI computes pairwise distances among $N$ chromosomes of length $L$, $\mathcal{O}(N^2 L)$. 
Genetic variation operators add $\mathcal{O}(N L)$. 
MADTS performs surrogate updates and acquisition over bounded archives, $\mathcal{O}(T_{LS} N)$. 
The Coordinator must also decode chromosomes and train MGNNs, incurring $C_{\text{train}}$, which dominates the runtime. 
Thus, the per-generation complexity is $\mathcal{O}(N^2 L + T_{LS} N + C_{\text{train}})$, and across $T$ generations it becomes $\mathcal{O}(T(N^2 L + T_{LS} N + C_{\text{train}}))$. 
In practice, $C_{\text{train}} \gg N^2 L, T_{LS}N$, so runtime is governed by model training and validation, 
while MADTS and SPDI \emph{cooperatively} enhance efficiency and stabilize the GA search dynamics.

\section{Experiments and Analysis}
\label{experiment and analysis}

\subsection{Experimental Settings}
\label{sec:exp_settings}

Following \cite{yin2022knowledge}, we evaluate the proposed framework on \textbf{VulCE}\footnote{\url{https://github.com/happyResearcher/VulKG}}, a multimodal vulnerability knowledge graph integrating CVE entries, CVSS severity vectors, and textual vulnerability descriptions. 

The task is formulated as \textbf{binary node classification}: given a vulnerability node, the model predicts whether it is high-risk or low-risk. Performance is measured by the \textbf{F1-score}, which balances precision and recall. We report both the \textbf{mean} and the \textbf{best} test F1 over 10 independent runs, and additionally record \textbf{GPU-hours} to quantify search and retraining cost.

All experiments are conducted on a server with 2$\times$RTX~4090 (24GB) GPUs, using PyTorch~2.1, DGL~1.1.2, and CUDA~12.1. The optimizer is Adam with learning rate $1\!\times\!10^{-3}$. Each run adopts a population size of $N=20$ (Eq.~\eqref{eq:population}) for $T=30$ generations. Chromosomes are encoded with length $K=18$ (Sec.~\ref{sec:ga_formulation}). In the MACC split (Sec.~\ref{sec:macc}), each worker returns $E^{(m)}_t = E^{(\text{fus})}_t = 5$ elites per generation (Eq.~\eqref{eq:macc-merge}). Local proposals are guided by \emph{MADTS} (Sec.~\ref{sec:dts}), while global variation is controlled by \emph{SPDI} (Sec.~\ref{sec:SPDI}). After the search, the best chromosome $\mathbf{C}^*$ is decoded into $\mathcal{M}(\mathbf{C}^*)$, retrained on the union of training and validation sets, and finally evaluated on the test set. Random seeds are varied across runs, and all configurations and logs are released for reproducibility.

\subsection{Compared Methods}

We compare MACC-MGNAS against both handcrafted multimodal GNNs and GNAS approaches. 
For human-designed models, we include \textbf{MAGCN (grid search)}~\cite{yin2022knowledge}, which relies on exhaustive tuning, as well as two recent architectures—\textbf{C2RS}~\cite{shu2025c} and \textbf{MyGO}~\cite{zhang2025tokenization}—that target cross-modal consistency and modality tokenization, respectively.

For GNAS competitors, we evaluate four classical search paradigms adapted to the multimodal setting: \textbf{GA}, \textbf{PSO}~\cite{huang2023split}, \textbf{EDA}~\cite{li2023surrogate}, and \textbf{BO}~\cite{zhang2023surrogated}. 
We also include \textbf{DC-NAS}~\cite{liang2024dc}, a divide-and-conquer evolutionary framework that optimizes modality and fusion components separately, without joint global coordination.

\paragraph{Parameter settings (fairness).}
To ensure a fair comparison, all NAS baselines (GA, PSO, EDA, BO, DC-NAS) use the same search budget as ours: population size $N=20$ and $T=30$ generations. 
Operator choices follow standard defaults reported in the literature. 
For MAGCN, C2RS, and MyGO, we adopt the authors' implementations and hyperparameters. 
All final architectures—regardless of search method—are retrained under an identical protocol (Adam, lr=$1\mathrm{e}{-3}$, early stopping), matching Sec.~\ref{sec:exp_settings}.

\subsection{Evaluation of Effectiveness}

\textbf{RQ1: Can MACC-MGNAS discover higher-performing architectures than manual designs and GNAS methods?}

To answer RQ1, we compare MACC-MGNAS against both manually designed MGNNs and NAS-based baselines. 
Table~\ref{tab:effectiveness-efficiency} reports mean/best F1-scores and GPU-hours averaged over 10 runs.

Among GNAS methods, MACC-MGNAS achieves the highest mean F1 (\textbf{81.67\%}) and best F1 (\textbf{84.04\%}), 
with statistically significant gains over \emph{all} baselines ($p<0.001$, paired $t$-test). 
Compared with the strongest competitor, BO (78.53\% mean F1), our method improves by +3.1 points 
while consuming fewer GPU-hours (3.0 vs.\ 4.9). 
Evolutionary baselines such as GA and PSO converge more slowly and yield lower F1 despite comparable or lower costs.

For \textbf{manual designs}, MAGCN achieves 72.96\% mean F1 but requires 12.55 GPU-hours. 
The high GPU cost arises from an exhaustive grid search over its design options, 
which substantially increases evaluation overhead. 
Recent MGNNs such as C2RS and MyGO perform slightly better (76--77\% mean F1) 
but still lag behind NAS-based approaches, underscoring the difficulty of manually tuning multimodal architectures.

Finally, \textbf{DC-NAS}, which adopts a divide-and-conquer strategy, achieves 77.65\% mean F1 at 4.10 GPU-hours, 
significantly below our method. 
This confirms that simply separating modality and fusion modules is insufficient; 
hierarchical coordination and modality-aware optimization in MACC-MGNAS are key to improved performance.

In summary, MACC-MGNAS consistently surpasses handcrafted MGNNs and outperforms state-of-the-art NAS algorithms, 
with statistically significant improvements, demonstrating strong effectiveness on VulCE.

\begin{table}[ht]
\centering
\renewcommand{\arraystretch}{1.15}
\setlength{\tabcolsep}{6pt}
\caption{Evaluation of MACC-MGNAS against existing MGNN methods, including GGNAS and fixed-architecture approaches, in terms of classification performance (F1) and GPU-hours efficiency. All results are based on 10 independent trials. $\dagger$: Statistically significant improvement over all baselines ($p<0.001$, paired \textit{t}-test).}
\label{tab:effectiveness-efficiency}
\begin{tabular}{l|c|c|c}
\toprule
\textbf{Method} & \textbf{Mean F1 (\%)} & \textbf{Best F1 (\%)} & \textbf{GPU-hours} \\
\midrule
PSO~\cite{huang2023split}         & $68.33 \pm 7.07$ & 79.34 & \textbf{2.75} \\
EDA~\cite{li2023surrogate}        & $73.19 \pm 6.16$ & 82.46 & 3.23 \\
GA~\cite{wang2025abg}             & $75.94 \pm 4.78$ & 81.72 & 3.95 \\
BO~\cite{zhang2023surrogated}     & $78.53 \pm 3.98$ & 82.31 & 4.90 \\
MyGO~\cite{zhang2025tokenization} & $74.30 \pm 2.41$ & 76.88 & 5.90 \\
C2RS~\cite{shu2025c}              & $76.12 \pm 3.17$ & 79.45 & 6.20 \\
DC-NAS~\cite{liang2024dc}         & $77.65 \pm 2.93$ & 80.92 & 4.10 \\
MAGCN~\cite{yin2022knowledge}     & $72.96 \pm 1.51$ & 74.60 & 12.55 \\
\midrule
\textbf{Ours} & $\mathbf{81.67 \pm 1.84}^\dagger$ & \textbf{84.04} & 3.00 \\
\bottomrule
\end{tabular}
\end{table}

\subsection{Evaluation of Efficiency}

\textbf{RQ2: Can MACC-MGNAS reduce computational cost while maintaining competitive or superior performance?}

\begin{figure}[ht]
\centering
\includegraphics[width=1\linewidth]{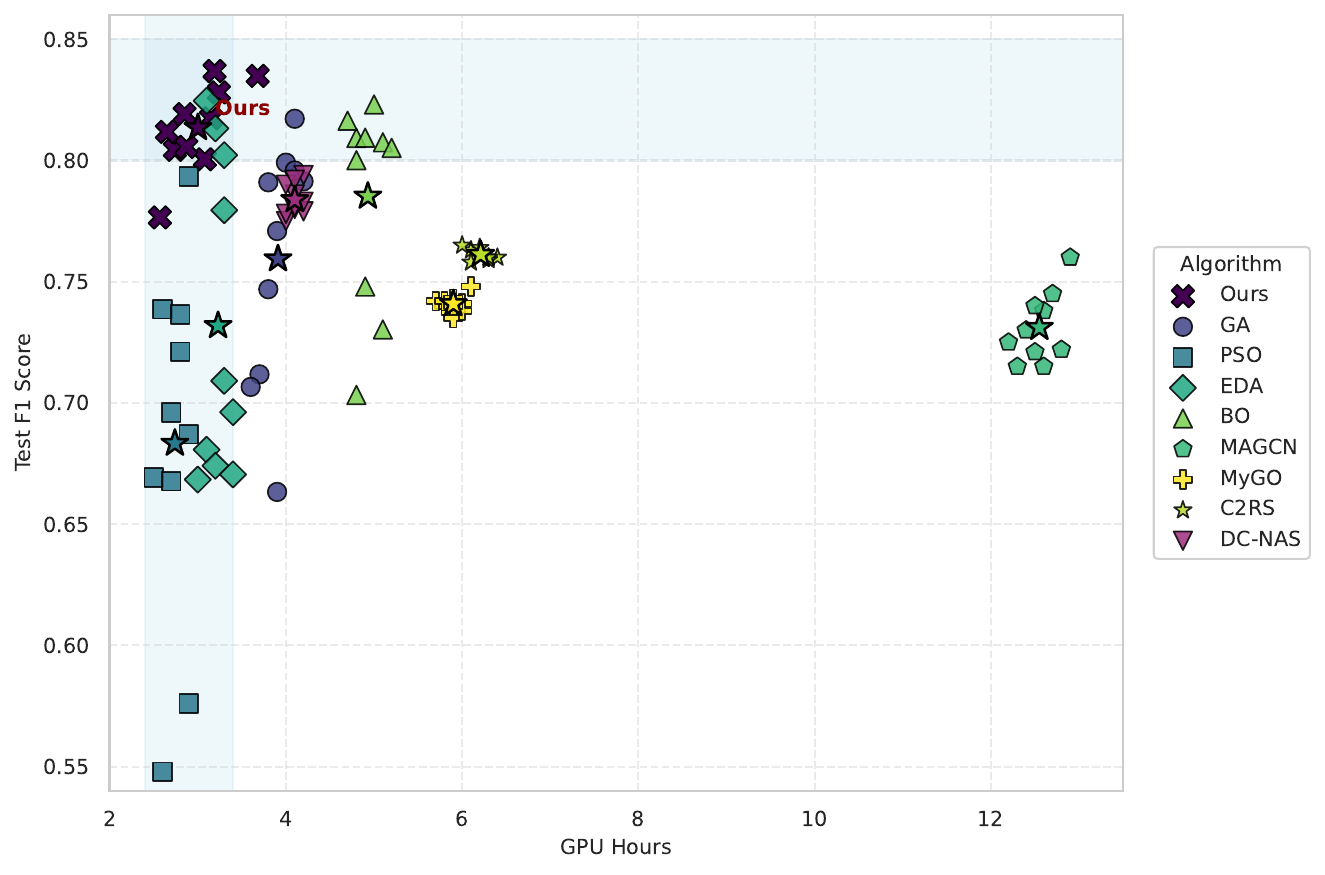}
\caption{Trade-off between efficiency and predictive accuracy across all compared methods.
Each point denotes one independent run (10 runs per method).
The $x$-axis reports GPU-hours for search and retraining, and the $y$-axis reports the corresponding test F1-score.
Marker colors and shapes indicate different algorithms (see legend).
MACC-MGNAS (purple crosses) consistently occupies the upper-left region, achieving higher F1 with lower GPU-hours than classical GNAS baselines (GA, PSO, EDA, BO) and multimodal frameworks (DC-NAS, MAGCN, MyGO, C2RS).
Note that PSO attains the lowest cost but clusters at lower F1, illustrating a cost–accuracy trade-off.
Overall, the distribution highlights the advantage of modality-aware decomposition in reducing cost without sacrificing accuracy.}
\label{fig:f1-vs-gpu}
\end{figure}

To answer RQ2, we assess efficiency in GPU-hours during search and retraining. Table~\ref{tab:effectiveness-efficiency} reports the average cost, and Fig.~\ref{fig:f1-vs-gpu} visualizes the accuracy–efficiency trade-off.

Overall, MACC-MGNAS requires only \textbf{3.0 GPU-hours}, lower than most GNAS baselines. For example, Bayesian Optimization (BO)---the strongest conventional GNAS competitor---consumes 4.90 GPU-hours, and DC-NAS incurs 4.10 GPU-hours while yielding lower accuracy. \emph{PSO} attains the lowest cost at \textbf{2.75 GPU-hours}, but this comes with the lowest mean F1 (68.33\%); compared with PSO, MACC-MGNAS uses only $+0.25$ GPU-hours yet improves mean F1 by \textbf{+13.34} points (81.67\% vs.\ 68.33\%) and best F1 by \textbf{+4.70} (84.04\% vs.\ 79.34\%). Manual designs such as MAGCN are far less efficient (12.55 GPU-hours) due to exhaustive grid search.

As shown in Fig.~\ref{fig:f1-vs-gpu}, MACC-MGNAS consistently lies in the upper-left region, combining higher F1 with low GPU cost. This favorable position reflects the benefit of our modality-aware, decomposed search that narrows the design space and accelerates convergence without sacrificing accuracy.

These results demonstrate that MACC-MGNAS alleviates the computational burden of conventional and divide-and-conquer GNAS frameworks, offering a practical and scalable solution for multimodal GNN design.

\subsection{Evaluation of Convergence and Adaptivity}
\label{sec:convergence-adaptivity}

\textbf{RQ3: Does MACC\text{-}MGNAS sustain effective exploration and avoid premature convergence through its adaptive mechanism (SPDI)?}

\begin{figure}[ht]
\centering
\includegraphics[width=1\linewidth]{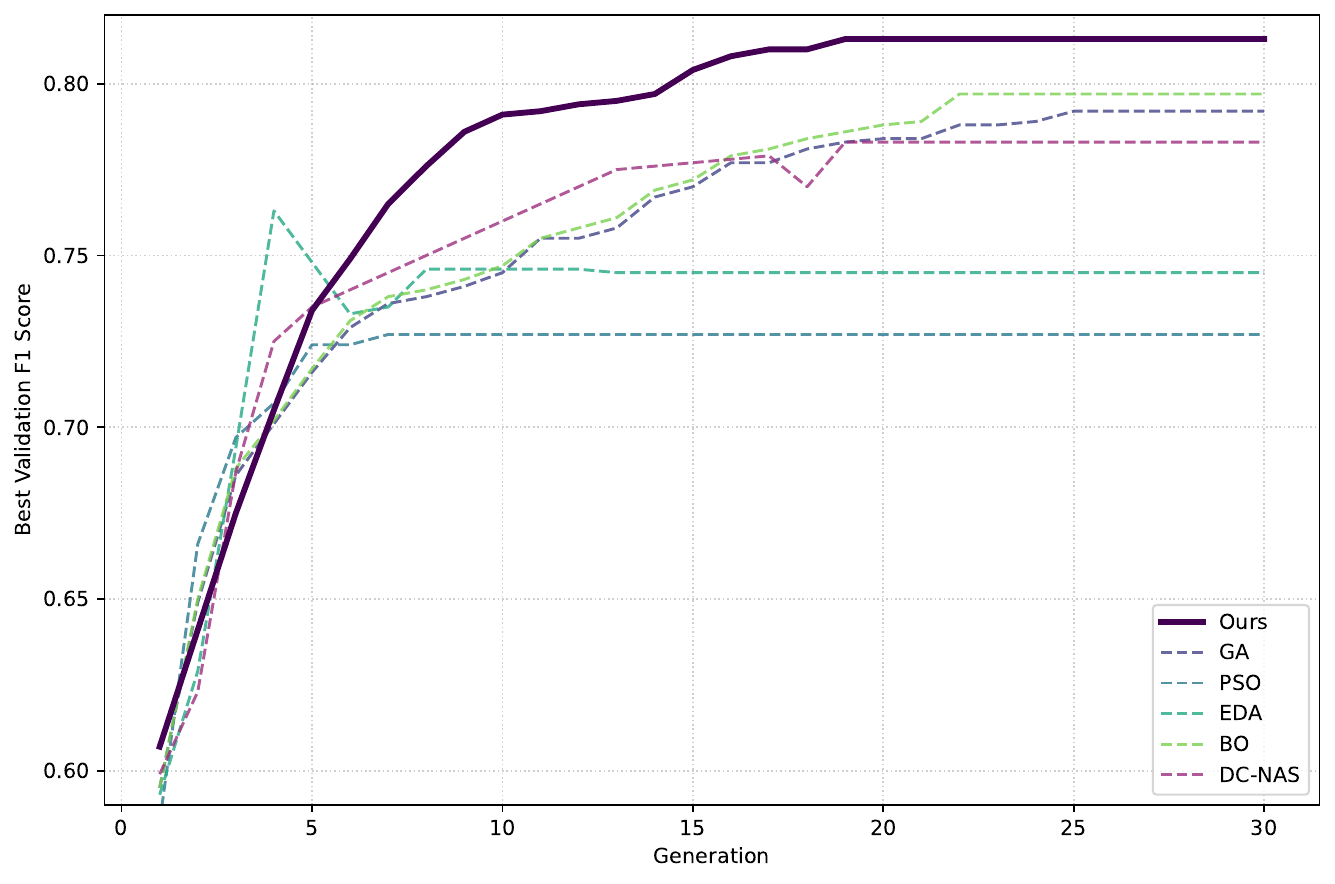}
\caption{Convergence comparison over 10 runs. 
MACC\text{-}MGNAS rises steeply early, reactivates exploration around Generation~17 through SPDI, and ultimately converges with the highest accuracy and lowest variance. The temporary drops in PSO and EDA curves are due to stochastic exploration generating weaker candidates before selection recovers. Other baselines either plateau early or converge more slowly.}
\label{fig:convergence_comparison}
\end{figure}

Figure~\ref{fig:convergence_comparison} presents validation F1 trajectories averaged over 10 runs.  
In the \emph{early stage} (Generations~1--10), MACC\text{-}MGNAS improves rapidly, reaching 0.81 by Generation~7, while GA, EDA, and PSO plateau below 0.78 and DC\text{-}NAS remains under 0.75.  

In the \emph{middle stage} (Generations~10--20), MACC\text{-}MGNAS exhibits a second lift around Generation~17, triggered by SPDI’s adaptive regulation of crossover and mutation. This diversity injection reactivates exploration without losing stability. Baselines lack such reactivation: GA, EDA, and PSO stagnate, BO progresses slowly, and DC\text{-}NAS stabilizes prematurely. Quantitatively, MACC\text{-}MGNAS reaches \textbf{0.81} by Gen.~7 (vs.\ GA/EDA/PSO $\le$\,0.78, DC\text{-}NAS $<\,0.75$), attains \textbf{0.83} by Gen.~20 (vs.\ BO $\approx$\,0.80 and others $\le$\,0.78), and finally converges near \textbf{0.83} with the smallest variance. Notably, the \textbf{temporary drops} in PSO and EDA (around Generations~5--8) are caused by their stochastic exploration strategies. Both algorithms occasionally generate weaker candidates through random perturbations, leading to short-lived decreases in population quality before selection pressure eliminates these suboptimal solutions. Quantitatively, these dips are small ($0.01$--$0.02$ F1, up to $0.03$ in rare cases) and typically recover within $2$--$3$ generations. This phenomenon underscores the instability of exploration when diversity is not adaptively regulated.

In the \emph{late stage} (Generations~20--30), MACC\text{-}MGNAS converges stably at 0.83 with the smallest variance. BO eventually approaches 0.82, while all other baselines saturate at lower levels.  

\begin{figure}[ht]
\centering
\includegraphics[width=1\linewidth]{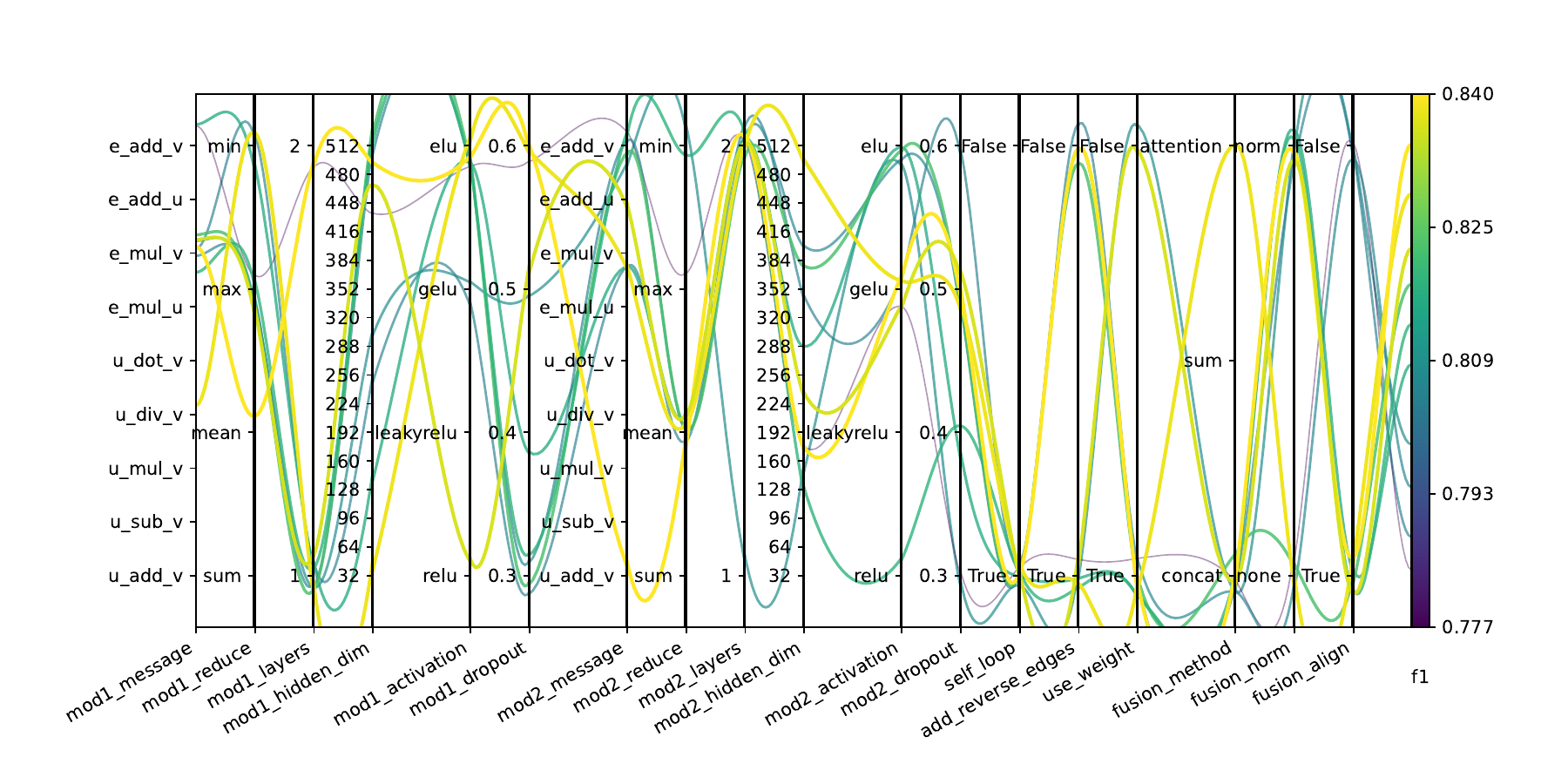}
\caption{Parallel coordinate plot of best architectures across 10 MACC-MGNAS runs. Lines denote architectures, with color encoding F1 scores. Consistent motifs such as multiplicative message operators and normalized fusion recur across trials, while residual diversity is preserved—evidence that SPDI consolidates effective patterns without collapsing exploration.}
\label{fig:trialall_parallel_plot}
\end{figure}

Figure~\ref{fig:trialall_parallel_plot} further examines the best architectures across 10 runs. Two consistent patterns emerge: (i) effective motifs (e.g., multiplicative message operators, GELU, normalized fusion) recur across runs, indicating reliable exploitation of strong designs; (ii) residual variation remains, confirming that SPDI sustains exploration rather than collapsing into a single structure. This balance of consolidation and diversity directly validates the adaptive role of SPDI.

In summary, MACC\text{-}MGNAS achieves faster and more stable convergence, while SPDI adaptively maintains exploration and prevents premature stagnation that commonly occurs in conventional NAS.

\subsection{Ablation Studies on MACC-MGNAS}
\label{sec:ablation}

We conduct ablation studies to evaluate the contributions of the three core components in MACC-MGNAS: MACC, MADTS, and SPDI. We compare four configurations: 
(1) \textit{w/o MACC}: a centralized GA without divide-and-conquer coordination, where Workers are disabled and the MADTS module is not applicable; 
(2) \textit{w/o MADTS}: a distributed MACC search with SPDI enabled but without surrogate modeling, requiring real evaluations for all candidates; 
(3) \textit{w/o SPDI}: a distributed MACC search with MADTS enabled but without adaptive diversity control, using fixed crossover and mutation rates; 
and (4) the full MACC-MGNAS framework integrating all three components. 

Table~\ref{tab:macc_ablation} reports mean and best F1 together with GPU-hours (averaged over 10 runs, including both search and retraining). It provides precise numerical comparisons and validates the effect of each design by contrasting ablated variants with the full model (via $\Delta$).

\paragraph{\textbf{Impact of MACC.}}  
The absence of divide-and-conquer coordination causes the most severe performance degradation. The \textit{w/o MACC} variant yields only 75.94\% mean F1, which is 5.8 points lower than \textit{w/o MADTS} (81.74\%). This confirms that modality-aware coordination is the primary driver of predictive accuracy, enabling modality-specific exploration while maintaining joint optimization.

\paragraph{\textbf{Impact of MADTS.}}  
Removing the surrogate has little effect on accuracy—\textit{w/o MADTS} achieves 81.74\% mean F1, close to the full model’s 81.67\%. However, efficiency degrades substantially: GPU-hours increase from 3.00 to 4.87. This demonstrates that MADTS mainly contributes to reducing computational cost by guiding sampling and reducing expensive real evaluations.

\paragraph{\textbf{Impact of SPDI.}}  
Disabling adaptive diversity control lowers mean F1 to 80.15\% and raises GPU-hours to 3.88. Moreover, the variance across trials increases. This indicates that SPDI plays a crucial role in stabilizing convergence and improving efficiency, by dynamically adjusting crossover and mutation rates to maintain an effective exploration–exploitation balance.

Collectively, the ablation results clarify complementary roles: MACC contributes the primary accuracy gains via modality-aware coordination, MADTS curtails computational cost by reducing full evaluations, and SPDI improves convergence stability through adaptive diversity control. Eliminating any single module compromises accuracy, efficiency, or robustness, underscoring that all three are required for the overall effectiveness of MACC-MGNAS.

\begin{table*}[ht]
\centering
\caption{Ablation study of the MACC-MGNAS framework. 
For \textit{w/o MACC}, $\Delta$ values are computed w.r.t. \textit{w/o DTS}, 
which corresponds to a centralized GA baseline. 
All other $\Delta$ values are computed w.r.t. the full MACC-MGNAS.}
\renewcommand{\arraystretch}{1.1}
\setlength{\tabcolsep}{5pt}
\begin{tabular}{l|c|c|c|c|c||c|c||c|c}
\toprule
\textbf{Method} & \textbf{MACC} & \textbf{DTS} & \textbf{SPDI} 
& \textbf{Mean F1 (\%)} & $\Delta$ 
& \textbf{Best F1 (\%)} & $\Delta$
& \textbf{GPU-hours} & $\Delta$ \\
\midrule
w/o MACC      
& \ding{55} & \ding{55} & \ding{51} 
& $75.94 \pm 4.78$ & $-5.80$ 
& 81.72 & $-1.66$
& 3.95 & $+0.32$ \\
w/o SPDI     
& \ding{51} & \ding{51} & \ding{55} 
& $80.15 \pm 2.38$ & $-1.52$
& 83.19 & $-0.85$
& 3.88 & $+0.88$ \\
w/o MADTG     
& \ding{51} & \ding{55} & \ding{51} 
& $\textbf{81.74} \pm \textbf{1.20}$ & $+0.07$
& 83.38 & $-0.66$
& 4.87 & $+1.87$ \\
\midrule
\textbf{MACC-MGNAS (Ours)}         
& \ding{51} & \ding{51} & \ding{51} 
& $\textbf{81.67} \pm \textbf{1.84}$ & --
& \textbf{84.04} & --
& \textbf{3.00} & -- \\
\bottomrule
\end{tabular}
\label{tab:macc_ablation}
\end{table*}

\subsection{Evaluation of Architecture Evolution and Interpretability}

\begin{figure}[ht]
\centering
\includegraphics[width=1\linewidth]{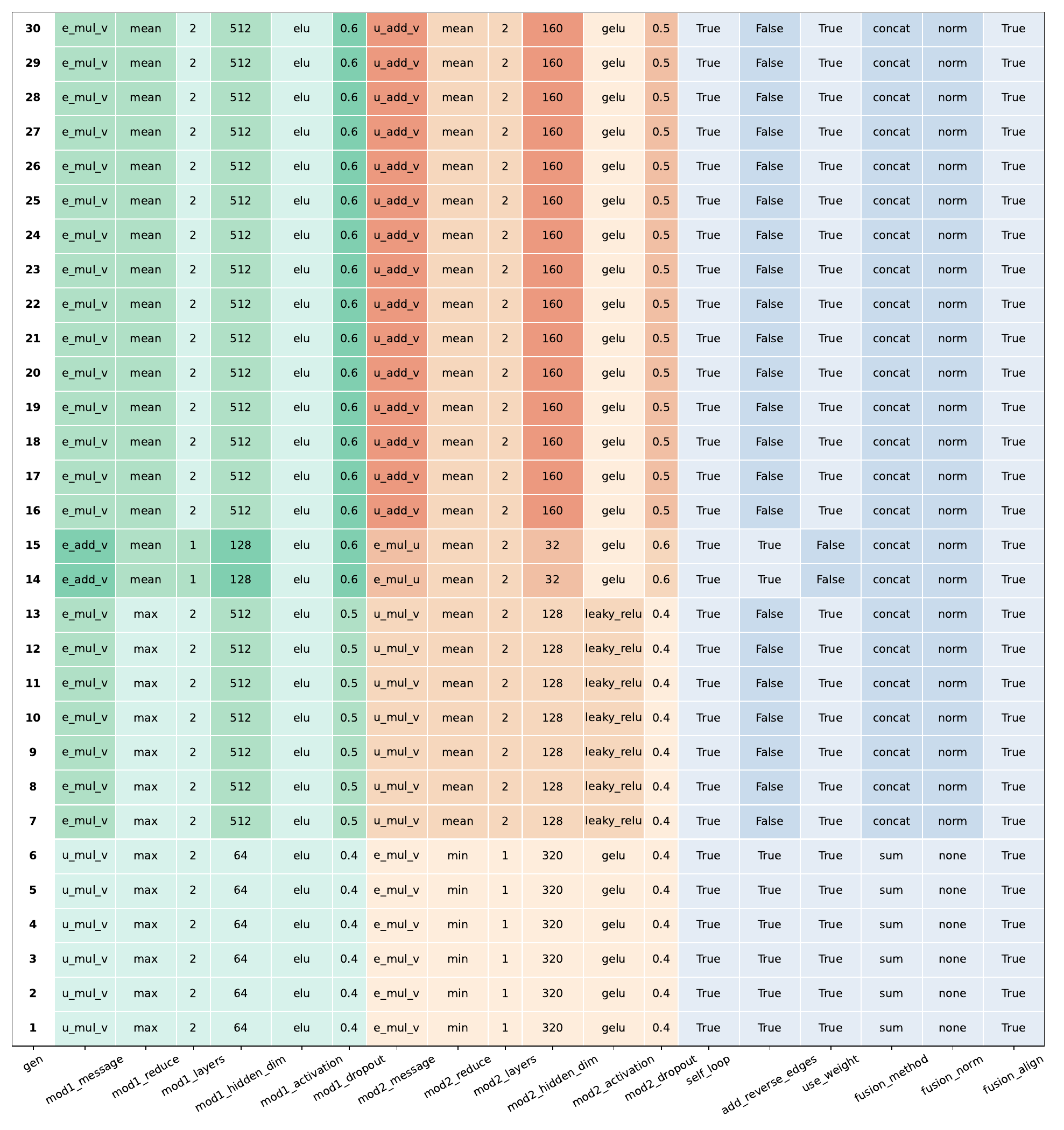}
\caption{Architecture evolution heatmap over 30 generations in a representative MACC-MGNAS run. 
Each row corresponds to a candidate architecture, with components grouped by modality (green = Modality~1, orange = Modality~2) and fusion (blue). 
Background shading distinguishes modalities, and the color bar indicates validation F1. 
The trajectory shows a gradual shift from additive to multiplicative message operators, increased hidden dimensions, and the emergence of normalized fusion modules (e.g., \texttt{concat+norm+align}), all of which align with performance improvements in later generations.}
\label{fig:masco_best_structures_heatmap}
\end{figure}

To further understand why MACC-MGNAS outperforms manual designs and conventional GNAS methods, we analyze the evolution of discovered architectures within a representative trial. Figure~\ref{fig:masco_best_structures_heatmap} illustrates the 30-generation trajectory, where each row denotes a candidate architecture and shading encodes validation F1. 

In the early stage, the population displays wide structural diversity with unstable scores, reflecting exploratory breadth without clear regularities. As evolution proceeds, systematic shifts emerge: message operators transition from additive (\texttt{e\_add\_v}) to multiplicative forms (\texttt{e\_mul\_v}, \texttt{e\_mul\_u}), hidden dimensions expand from 128 to 512 to capture richer representations, and fusion modules evolve toward normalized alignment strategies such as \texttt{concat+norm+align}. The appearance of these modules around Generation~16 coincides with a marked jump in validation F1, underscoring their critical role in enhancing predictive accuracy.

Taken together, these structural regularities reveal design principles distilled through data-driven search: (i) leverage multiplicative message interactions for expressive relational reasoning, (ii) allocate sufficient hidden capacity to avoid representational bottlenecks, and (iii) employ normalization and alignment in multimodal fusion to mitigate modality imbalance. Importantly, these principles extend beyond the immediate benchmark: they provide generalizable guidelines for initializing MGNN designs, pruning the search space in future NAS frameworks, and informing manual architectures in broader multimodal graph learning contexts. By uncovering such transferable insights, MACC-MGNAS contributes not only higher empirical performance, but also a deeper understanding of what constitutes effective multimodal GNN design.

\section{Conclusion}
\label{conclusion}

In this paper, we studied co-exploitation prediction on multimodal vulnerability knowledge graphs, a task hindered by the structural heterogeneity of MGNNs, the vast discrete design space, and the evaluation inefficiency of GNAS. We proposed \emph{MACC-MGNAS}. It integrates three key components: \emph{MACC} for jointly optimizing modality-specific and fusion gene blocks, \emph{MADTS} for efficient fitness estimation, and \emph{SPDI} for balancing exploration and exploitation. Extensive experiments on the VulCE benchmark demonstrate that \emph{MACC-MGNAS} achieves an F1-score of 81.67\% within 3 GPU-hours, surpassing the state-of-the-art by +8.7\% F1 while reducing computational cost by 27\%. Ablation and convergence analyses further confirm that \emph{MACC} effectively models modality heterogeneity, \emph{MADTS} substantially improves efficiency, and \emph{SPDI} enhances stability and convergence. Moreover, architecture evolution analysis distills interpretable design principles such as multiplicative interactions, larger hidden dimensions, and normalized fusion.  

Despite these advances, \emph{MACC-MGNAS} still faces three limitations: (i) evolutionary NAS remains costly on very large graphs despite the surrogate acceleration; (ii) the use of a predefined chromosome length restricts the openness of the search space and may overlook novel designs; and (iii) current experiments are limited to static graphs, whereas real-world vulnerability data are dynamic and evolving. Future work will address these issues by further improving scalability, enabling data-driven space expansion, and extending the algorithm to continual and online learning.

\bibliographystyle{IEEEtran}
\bibliography{p2_submit}

\end{document}